# Mutual Information calculation on different appearances

Jiecheng Liao   Junhao Lu   Jeff Ji   Jiacheng He

*Abstract*—Mutual information has many applications in image alignment and matching, mainly due to its ability to measure the statistical dependence between two images, even if the two images are from different modalities (e.g., CT and MRI). It considers not only the pixel intensities of the images but also the spatial relationships between the pixels. In this project, we apply the mutual information formula to image matching, where image A is the moving object and image B is the target object and calculate the mutual information between them to evaluate the similarity between the images. For comparison, we also used entropy and information-gain methods to test the dependency of the images. We also investigated the effect of different environments on the mutual information of the same image and used experiments and plots to demonstrate.

## I. Introduction

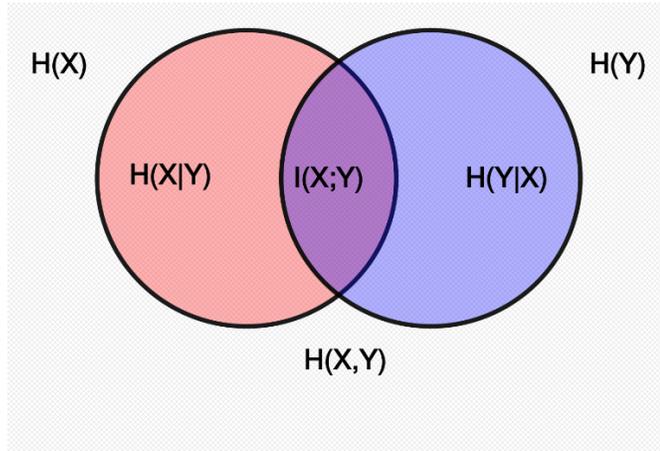

$figure\ 1$

Mutual information (MI) between two random variables is a measure of the mutual reliance between the two variables in information theory and probability theory. A Venn diagram illustrates the correlations, both additive and subtractive, between several information measures and the correlated variables $X$ and $Y$. The joint entropy $H(X,Y)$ is the area that is enclosed by either circle. The conditional entropy H $(X|Y)$ is shown by the red circle on the left, while the individual entropy H(X) is represented by the violet circle. The blue circle on the right is $H(Y|X)$, while the violet circle is $H(X|Y)$. In terms of mutual information, the violet is $I(X,Y)$. If the random variables $X$ and $Y$ are independent, the mutual information is 0.

This concept, akin to entropy, can be applied in various fields, including image analysis. This report proposes an innovative application of mutual information to assess the similarity in appearance between different individuals. By capturing images of ten pairs of individuals using a cell phone camera, we compute the mutual information between these pairs. The aim is to determine the degree of resemblance between each pair based on the mutual information calculated. This approach offers a novel perspective on image comparison and opens new possibilities for its application in fields such as biometrics, social networking, and even entertainment. We also compare them with images that have used entropy and information-gain methods. The fact that different environmental variables have a large effect on mutual information is also explored in the follow-up. We visualize the results of our experiments in matrices, graphs, etc. The following sections will delve into the methodology, results, and implications of this intriguing study.

## II. Formula Analysis

$$MI(A,B) = \sum_{i_a, j_b} p_{AB}(i_a, j_b) \log_2\left(\frac{p_{AB}(i_a, j_b)}{p_A(i_a) p_B(j_b)}\right)$$

image A (moving)
image B (target)
intensity level in image A

$figure\ 2$

In the formula, **MI(A, B)** is mutual information between image A and image B. $(i_a), (j_b)$ represent intensity levels (pixel value) in images A and B. $p_{AB}(i_a, j_b)$ indicates a joint probability mass function at where pixels are $i_a$ and $j_b$ in two images. This function gives the pixel intensity in image A as $i_a$ and the pixel intensity in image B is $j_b$ The probability of simultaneous occurrence. $p_A(i_a), p_B(j_b)$ mean marginal probabilistic mass functions of the intensities of the pixels in images A and B, respectively, $i_a, j_b$. This function respectively gives the probability that the pixel intensity in image A is $i_a$ and the pixel intensity in image B is $j_b$.

The main function of this formula is to measure the similarity of two images by comparing their pixel intensity distributions.

.

If there is a high correlation between the pixel intensities of the two images, the greater the joint probability mass function $i_a, j_b$, the smaller the product that $p_A(i_a)$ and $p_B(j_b)$, then their mutual information will be high.

### III. EXPERIMENT

#### A. Image similarity calculation by mutual information

Core step:
1. Pre-processing:
- Convert RGB images to grayscale.
- Address any size discrepancies between the images.
- Adjust the intensity settings of the images.
2. PDF Calculation:
- Calculate the probability density function (pdf) for the two images separately.
- We are using a 2D histogram to calculate the joint pdf of 2 images.
3. Mutual Information Calculation:
- calculate MI according to the formula(figure 2) by using the pdf we calculated before.

Here is the pseudo-code of our calculations:

1. Import necessary libraries: os, numpy, OpenCV, PIL Image, matplotlib.pyplot
2. Define function compute_probability_distribution(image):
   a. Flatten the image to a 1D array of intensity values.
   b. Calculate the probability distribution (histogram) of these intensity values.
   c. Return the histogram.
3. Define function mutual_information(imageA, imageB):
   a. Compute the probability distribution of each image using compute_probability_distribution.
   b. Compute the joint probability distribution of both images.
   c. Initialize mutual information (mi) to 0.
   d. Iterate over all intensity levels:
     i. If joint probability and individual probabilities are greater than 0, update mi.
   e. Return the mutual information.
4. Define a variable size with a value of (256, 256).
5. Define function load_image_with_opencv(file_path, size=None):
   a. Read the image file using OpenCV.
   b. Convert the image from BGR to grayscale.
   c. If a size is provided, resize the image.
   d. Return the processed image.

We then do the calculations based on the 5 pictures to form 25 pairs of appearances:

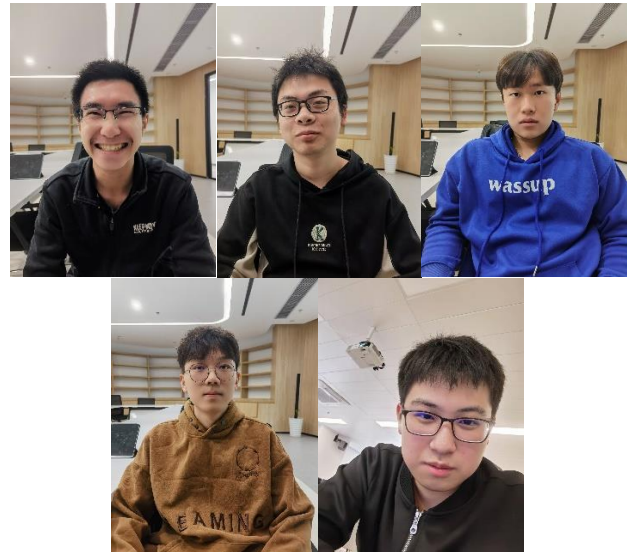

*figure* 3

After doing the MI calculations, we get all the data,

*figure* 4

Then we form it into a matrix, then we can get a regularity:

*Higher MI, higher similarity*

MI matrix:

*figure* 5

We can see the regularity more clearly by drawing the following plot:

X is the difference images, and y is the calculated MI; We can see that only the calculation between the images itself has a high MI.

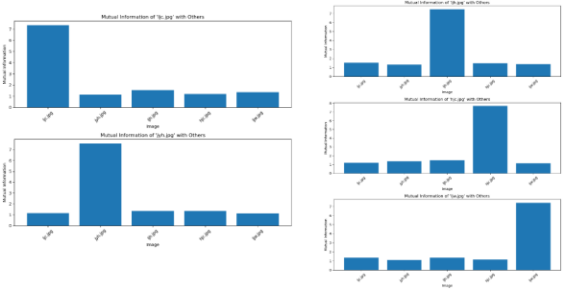

*figure* 6

To verify whether it can judge the same appearances in different environments, we experiment with the 2 following images:

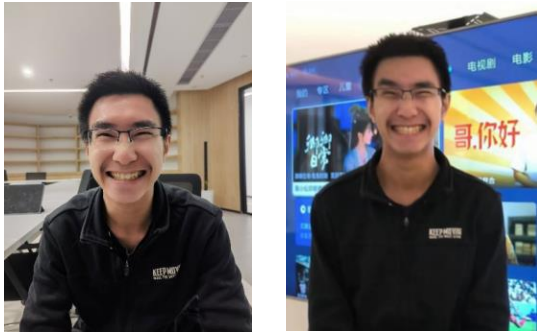

*figure* 7

We can see that it doesn't work well, because the information is lost during the computation,

```
The mutual information is: 0.5757517978089428
```

We also use the MI calculation method on *sklearn*, to show that it also gets a similar result, to troubleshoot our algorithms and prove our above statement is true.

Hence, using MI to compare the appearance of different people in different backgrounds directly is awkward.

### B. *Image similarity calculation by Entropy and Information Gain*

In this section, we calculated the entropy and information gain of the image. Firstly, we calculated the entropy of the RGB image. We selected five images, calculated the entropy of each image, and then merged them with the other four images to calculate the entropy after merging.

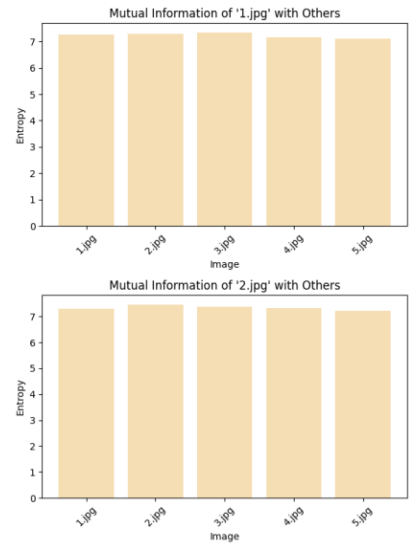

*figure* 8

Then we calculated the information gain of RGB images to measure the contribution of a certain feature or attribute in the image to the overall information content.

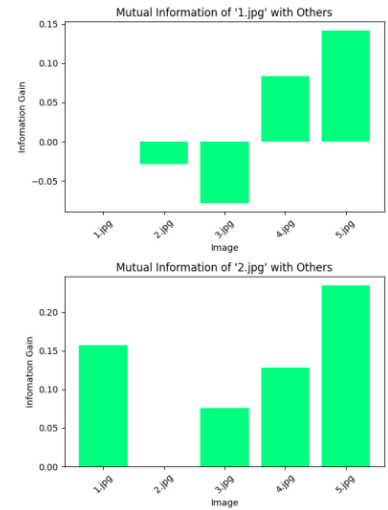

*figure* 9

By calculating the entropy and information gain of the image, the effectiveness of image merging can be evaluated. The entropy value of the merged image has changed relative to the entropy value of the original image, which can be used to measure the degree of change in image information by the merging algorithm. By comparing the Information Gain before and after merging, we can understand which information has been enhanced or weakened during the merging process.

## IV. ALGORITHM EVALUATION

**Mutual information:**

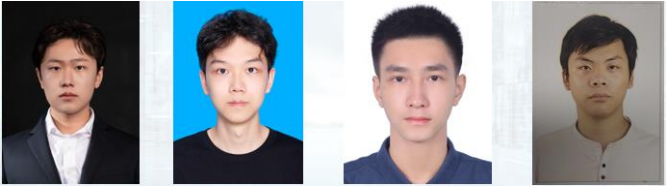
figure 10

To reevaluate the two algorithms more precisely, we have opted for headshot photographs with consistent shooting range and facial expressions in the background. They represent our selected subjects for the photos.

```
Mutual Information Matrix:
[[5.69621174 0.70446835 1.23264485 0.85737603]
 [0.70446835 6.89774482 0.93536942 0.85543791]
 [1.23264485 0.93536942 5.90605064 0.97274356]
 [0.85737603 0.85543791 0.97274356 6.49055315]]
```
figure 11

The first aspect involves the re-evaluation of mutual information. In this analysis, we observe similarities to the previous group, except for a relatively high self-evaluation, while the evaluations for others remain low. This suggests that overall similarity remains relatively modest.

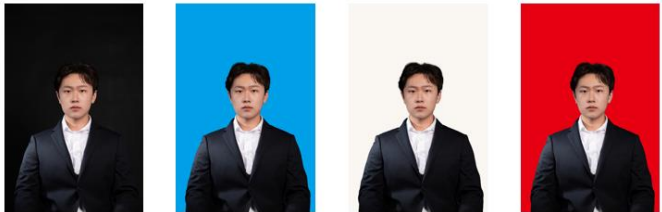
figure 12

However, if we introduce a change by analyzing a headshot of the same individual against a backdrop with a different color, we anticipate that the similarity should become more approximate.

```
[[4.68501157 4.57561955 4.49496789 2.40920648]
 [4.57561955 4.70065081 4.51114144 2.41571857]
 [4.49496789 4.51114144 4.623184   2.39824021]
 [2.40920648 2.41571857 2.39824021 5.57878913]]
```
figure 13

Contrary to our expectations, the table data indicates that the similarity between the red background panel and the others is not high. This outcome challenges our initial assumption that all similarities would align, suggesting a deviation from our anticipated results.

```
[[2.88890358e+00 1.79839134e-02 4.07123191e-04 4.07123191e-04]
 [1.79839134e-02 2.28390910e+00 1.58496250e+00 1.58496250e+00]
 [4.07123191e-04 1.58496250e+00 1.58496250e+00 1.58496250e+00]
 [4.07123191e-04 1.58496250e+00 1.58496250e+00 1.58496250e+00]]
```
The mutual information between '61.jpg' and '61.jpg' is: 2.888903584644827
The mutual information between '61.jpg' and '62.jpg' is: 0.0179839134115203
The mutual information between '61.jpg' and '63.jpg' is: 0.000407123191081442
The mutual information between '61.jpg' and '64.jpg' is: 0.000407123191081442
The mutual information between '62.jpg' and '61.jpg' is: 0.0179839134115204
The mutual information between '62.jpg' and '62.jpg' is: 2.283909101565316
The mutual information between '62.jpg' and '63.jpg' is: 1.5849625007211565
The mutual information between '62.jpg' and '64.jpg' is: 1.5849625007211565
The mutual information between '63.jpg' and '61.jpg' is: 0.000407123191081442
The mutual information between '63.jpg' and '62.jpg' is: 1.5849625007211565
The mutual information between '63.jpg' and '63.jpg' is: 1.5849625007211559
The mutual information between '63.jpg' and '64.jpg' is: 0.000407123191081442
The mutual information between '64.jpg' and '61.jpg' is: 0.000407123191081442
The mutual information between '64.jpg' and '62.jpg' is: 1.5849625007211563
The mutual information between '64.jpg' and '63.jpg' is: 1.5849625007211559
The mutual information between '64.jpg' and '64.jpg' is: 1.5849625007211559

figure 14

Hence, to delve deeper into the algorithm's proficiency in image recognition based on color intensity, we exclusively employ the background color for recognition in this round of testing. The obtained results validate our hypothesis, reinforcing the notion that the algorithm's performance is indeed contingent on color intensity. However, it's worth noting that the current formula exhibits a significant drawback, as it assesses different pixels in a suboptimal manner.

**Entropy and information Gain:**

The value of the entropy matrix reflects the similarity or difference between each pair of images. Images with lower values may be similar in some way, while higher values may indicate larger differences.

The information gain matrix provides the overall uncertainty change after merging the images. Negative values indicate that the combined image becomes more regular or predictable overall.

```
Entropy Matrix:
[[5.41246128 6.62685156 5.59468174 6.47840643]
 [6.62685156 6.6458416  6.23047161 6.22228718]
 [5.59468174 6.23047161 5.03680611 6.24137306]
 [6.47840643 6.22228718 6.24137306 6.32437801]]

Info_gain Matrix:
[[ 0.         -1.21439028 -0.18222046 -1.06594515]
 [ 0.01899004  0.          0.41536999  0.42355442]
 [-0.55787563 -1.1936655   0.         -1.20456696]
 [-0.15402842  0.10209084  0.08300495  0.        ]]

[[7.25762224 7.17424107 7.33552122 7.28614998]
 [7.17424107 7.33342123 7.40551615 7.31565285]
 [7.33552122 7.40551615 7.52035522 7.36772823]
 [7.28614998 7.31565285 7.36772823 7.44334126]]
Info_gain Matrix:
[[ 0.          0.08338118 -0.07789898 -0.02852774]
 [ 0.15918016  0.         -0.07209492  0.01776838]
 [ 0.184834    0.11483908  0.          0.15262699]
 [ 0.15719128  0.12768841  0.07561302  0.        ]]
```
figure 15

For entropy and information gain, we adopt a consistent analysis and comparison approach, following the same order as the previous four individuals' headshots. While the differences in results are not notably significant, some values have shifted into the negative range, indicating an elevated level of similarity. This shift may be attributed to variations in the background, suggesting its potential influence on the outcomes.

For Mug shot:

```
Entropy Matrix:
[[3.65646887 3.71341252 3.71833897 5.05310059]
 [3.71341252 3.70640326 3.69884205 5.40979004]
 [3.71833897 3.69884205 3.70133305 5.43182898]
 [5.05310059 5.40979004 5.43182898 5.4932723 ]]
Info_gain Matrix:
[[ 0.         -0.05694366 -0.0618701  -1.39663172]
 [-0.00700927  0.          0.00756121 -1.70338678]
 [-0.01700592  0.002491    0.         -1.73049593]
 [ 0.44017172  0.08348227  0.06144333  0.        ]]
```

*figure* 16

For Pure background:

```
Entropy Matrix:
[[ 2.893538    1.86685181  1.72632074  1.86508536]
 [ 1.86685181  0.04541467  0.04541467 -0.        ]
 [ 1.72632074  0.04541467 -0.         -0.        ]
 [ 1.86508536 -0.         -0.         -0.        ]]
Info_gain Matrix:
[[ 0.          1.02668619  1.16721725  1.02845263]
 [-1.82143712  0.          0.          0.04541467]
 [-1.72632074 -0.04541467  0.          0.        ]
 [-1.86508536  0.          0.          0.        ]]
```

*figure* 17

By employing the same processing method as before, it is evident that the color remains red, mirroring the previous outcome. This indicates that the results from both algorithms are comparable, highlighting that neither of them excels in accurately recognizing similar images.

## V. CONCLUSION

The study demonstrates the utility of mutual information (MI) in image matching, showcasing its effectiveness in assessing similarity based on pixel intensities and spatial relationships. However, the limitations of MI become apparent when comparing appearances in diverse environments, indicating potential information loss during computation. The introduction of entropy and information gain provides alternative methods for image merging analysis. The evaluation suggests that the algorithm's performance is influenced by color intensity, with MI showing comparable results to entropy and information gain in recognizing similar images. Further research is warranted to address the identified limitations and enhance the algorithm's robustness across varied scenarios.